\documentclass[conference,compsoc]{IEEEtran}

\usepackage{hyperref, url}
\usepackage{comment}
\usepackage{amssymb}
\usepackage{tikz}

\ifCLASSOPTIONcompsoc
  \usepackage[nocompress]{cite}
\else
  \usepackage{cite}
\fi

\begin{document}

\title{Dataflow Matrix Machines as a Model of Computations with Linear Streams}

\author{\IEEEauthorblockN{Michael Bukatin}
\IEEEauthorblockA{HERE North  America LLC\\
Burlington, MA, USA\\
Email: bukatin@cs.brandeis.edu}
\and
\IEEEauthorblockN{Jon Anthony}
\IEEEauthorblockA{Boston College\\
Chestnut Hill, MA, USA\\
Email: jsa.aerial@gmail.com}}

\maketitle

\begin{abstract}
We overview dataflow matrix machines as a Turing complete generalization of recurrent neural networks
and as a programming platform. We describe vector space of finite prefix trees with numerical leaves which allows us to
combine expressive power of dataflow matrix machines with simplicity of traditional recurrent neural networks.
\end{abstract}

\IEEEpeerreviewmaketitle

\section{Introduction}

When one considers a Turing complete generalization of recurrent neural networks (RNNs),
four groups of questions arise naturally: a) what is the mechanism providing access
to unbounded memory; b) what is the pragmatic power of the available primitives, and
is the resulting platform suitable for crafting software manually, rather than only
serving as compilation and machine learning target; c) what are self-referential
(and self-modification) mechanisms if any; d) what are the implications for
machine learning. 

In Section~\ref{sec:overview} we overview dataflow matrix machines, a generalization
of RNNs based on arbitrary linear streams, neurons of arbitrary nonnegative input and output arity,
a novel model of unbounded memory, and well-developed self-referential facilities,
following~\cite{BMR2, BMR3, BMR4}.

Dataflow matrix machines are much closer to being a general-purpose programming
platform than RNNs, while retaining the key property of RNNs that large classes of programs can be
parametrized by matrices of numbers, and therefore synthesizing appropriate matrices is sufficient
to synthesize programs.

In Section~\ref{sec:core} we describe the formalism based on the vector space of finite prefix trees with numerical leaves
which is used in our current  Clojure implementation of the core primitives of dataflow matrix machines~\cite{DMM}.

The concluding Section~\ref{sec:conclusion} discusses some of possible uses of dataflow matrix machines
in machine learning.

\section{Dataflow Matrix Machines: an Overview}\label{sec:overview}

\subsection{Countable-sized Nets with Finite Active Part}
One popular approach to providing Turing complete generalizations of RNNs
with unbounded memory is to use an RNN as a controller to a Turing machine
tape or another model of external memory~\cite{McCullochPitts},~\cite{GravesWayneDanihelka},~\cite{WestonChopraBordes}.

Another approach is to allow reals of unlimited precision, in effect using
a binary expansion of a real number as a tape of a Turing machine~\cite{SiegelmannSontag}.

Dataflow matrix machines take a different approach. One considers a countable-sized
RNN, and therefore a countable matrix of connectivity weights, but with
a condition that only finite number of those weights are non-zero at any given
moment of time. 

At any given moment of time, only those neurons are active which have at least
one non-zero connectivity weight associated with them. Therefore only a finite
part of the network is active at any given time.

Memory and network capacity can be dynamically added 
by gradually making more weights to become non-zero~\cite{BMR2}.

\subsection{Dataflow Matrix Machines as a Generalization of Recurrent Neural Networks}

The essence of neural models of computations is to interleave generally non-linear,
but relatively local computations performed by the activation functions built into neurons,
and linear, but potentially quite global computations recomputing neuron inputs from
the outputs of various neurons.

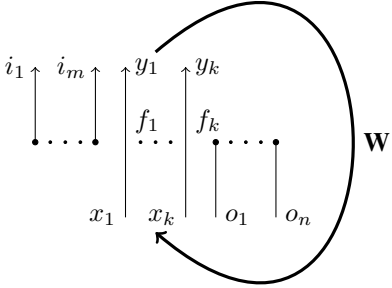
\begin{figure}[h]
\begin{tikzpicture}
   \clip (-2.0, -2.0) rectangle (4.0, 2.0);
   \draw [->] (-0.8, 0) -- (-0.8, 1) node [left] {$i_m$};
   \draw [->] (-1.6, 0) -- (-1.6, 1) node [left] {$i_1$};
   \filldraw (-1.2,0) circle [radius=0.5pt]
                 (-1.4,0)  circle [radius=0.5pt]
                 (-1.0, 0) circle [radius=0.5pt]
                 (-0.8, 0) circle [radius=1pt]
                 (-1.6, 0) circle [radius=1pt];

   \draw [->] (-0.4, -1) node [left] {$x_1$} -- (-0.4, 1) node [midway, above right] {$f_1$} node [right] {$y_1$};
   \draw [->] (0.4, -1)  node [left] {$x_k$} -- (0.4, 1) node [midway, above right] {$f_k$} node [right] {$y_k$};
   \filldraw (0,0) circle [radius=0.5pt]
                 (-0.2,0)  circle [radius=0.5pt]
                 (0.2, 0) circle [radius=0.5pt];

   \draw (0.8, -1) node [right] {$o_1$} -- (0.8, 0);
   \draw (1.6, -1) node [right] {$o_n$} -- (1.6, 0);
   \filldraw (1.2,0) circle [radius=0.5pt]
                 (1.4,0)  circle [radius=0.5pt]
                 (1.0, 0) circle [radius=0.5pt]
                 (0.8, 0) circle [radius=1pt]
                 (1.6, 0) circle [radius=1pt]; 

  \draw [->, very thick] (0, 1.2) .. controls (3.5, 4.2) and (3.5, -4.2) .. (0, -1.2)  node [midway, right] {{\bf W}};

\end{tikzpicture}
\caption{``Two-stroke engine" for an RNN. ``Down movement": 
$(x_1^{t+1}, \ldots, x_k^{t+1}, o_1^{t+1}, \ldots, o_n^{t+1})^{\top}=$ {\bf W}$ \cdot (y_1^{t}, \ldots, y_k^{t}, i_1^{t}, \ldots, i_n^{t})^{\top}$.
``Up movement": $y_1^{t+1} = f_1(x_1^{t+1}), \ldots, y_k^{t+1} = f_k(x_k^{t+1})$.}
\label{fig:rnn}
\end{figure}

The metaphor of a ``two-stroke engine" is applicable to traditional RNNs. 
On the ``up movement", the activation
functions built into neurons are applied to the inputs of the neurons and produce
the next values of the output streams of the neurons. On the ``down movement", the
matrix of connectivity weights ({\em network matrix}) is applied to the (concatenation of the) vector of neuron outputs (and the vector
of network inputs) and produces the (concatenation of the)
vector of the next values of the input streams of all neurons (and the vector of network outputs). This ``two-stroke cycle"
is repeated indefinitely (Fig.~\ref{fig:rnn}).

Dataflow matrix machines (DMMs) attempt to generalize RNNs as much as possible, while
preserving this structure of the ``two-stroke engine". In particular, the key element to be preserved
is the ability to apply the matrix of connectivity weights to the collection of neuron
outputs, hence the notion of linear combination must be defined.

RNNs work with streams of numbers. DMMs work
with streams of approximate representations of arbitrary vectors ({\em linear streams}).
One considers a finite or countable collection of {\em kinds of linear
streams}. With every kind of linear stream $k$, one associates a vector space $V_k$ and
a way to compute an approximate representation of vector $\alpha_1 v_{1,k} + \ldots + \alpha_n v_{n,k}$ from
approximate representations of vectors $v_{1,k}, \ldots, v_{n,k}$.

A {\em neuron type} has a non-negative integer input arity $I$, a non-negative integer output
arity $J$, kinds of linear streams $i_1, \ldots, i_I$ and $j_1, \ldots, j_J$ associated with
neuron inputs and outputs, and an activation function associated with this neuron type.
In the simplest version, the activation function
maps $V_{i_1} \times \ldots \times V_{i_I}$ to $V_{j_1} \times \ldots \times V_{j_J}$.
In reality, one needs to consider the fact that one works not with vectors, but with
their approximate representations, that activation functions might be stochastic, etc.
In particular, Appendix~\ref{sec:linear} discusses how linear streams of probabilistic samples fit this framework.

One considers a finite or countable collection of neuron types, and a countable set of
neurons of each type. For each output of each neuron, and for each input of each neuron,
the network matrix has a weight coefficient connecting them. At any given time,
only finite number of those coefficients can be non-zero, and moreover only weight
coefficients connecting outputs and inputs which have the same kind of linear streams
associated with them are allowed to be non-zero (a type correctness condition). See Fig.~\ref{fig:dmmold}
in Appendix~\ref{sec:twostrokedmmold}.

\subsection{Pragmatic Power of Dataflow Matrix Machines as a Programming Platform}

The pragmatic power of dataflow matrix machines is considerably higher than the
pragmatic power of vanilla RNNs~\cite{BMR3}. The ability to handle streams of
sparse representations of arrays is instrumental for the ability to implement
various algorithms based on hash maps and similar structures without extra runtime and
memory overhead.

Neurons with linear activation functions such as identity allow us to implement memory
primitives such as accumulators, leaky accumulators, etc.

The ability to have multiple inputs allows us to have multiplicative neurons implementing
mechanisms for gating (``multiplicative masks"), which serve as fuzzy conditionals and
can be used to attenuate and redirect flows of data in the network~\cite{Pollack}. Multiplicative neurons are
implicitly present in modern recurrent neural network architectures such as LSTM and Gated
Recurrent Unit networks (Appendix C of~\cite{BMR4}).

The sparseness structure of the network matrix can be used to sculpt the layered structure
and other topological features of the network, and multiplicative neurons can be used to orchestrate multilayered computations
by silencing particular layers at appropriate moments of time.

The fact that streams of samples can be used to represent streams of probability distributions and
signed measures allows us to incorporate certain streams of non-vector objects without explicitly
embedding those objects into vector spaces.

The ability to handle streams of arbitrary vectors and to have arbitrary input and output arities
considerably increases the ability to structure and modularize the resulting networks and programs.

\subsection{Self-referential Mechanism}

There is a history of research studies suggesting that it might be fruitful for a neural network to
be able to reference and update its own weights~\cite{Schmidhuber}. However, doing this with
standard neural networks based on scalar streams is difficult. One has to update the network
matrix on per-element basis, and one needs to encode the location of matrix elements
(row and column indices) within real numbers. This often results in rather complicated and
fragile structures, highly sensitive to small changes of parameters.

When networks can process arbitrary linear streams, self-referential mechanisms become much
easier. One simply incorporates neurons processing streams of matrices, requiring those matrices
to have shapes appropriate for network matrices in a given context. Then one can dedicate
a particular neuron {\tt Self} and use its latest output as the network matrix~\cite{BMR2}.

The {\tt Self} neuron is typically implemented as an accumulator, allowing it to take
incremental updates from the other matrix outputting neurons in the network.

Appendix~\ref{sec:selfref} presents a self-contained simple example of a self-referential
dynamical system, where our basic network matrix update mechanism together with a few
constant update matrices produce a wave pattern of connectivity weights dynamically propagating within the network
matrix.

The updating neurons can access the network matrix via their inputs, which allows them
to perform sophisticated computations. For example, one can have updating neurons
creating deep copies of network subgraphs and use those to build pseudo-fractal structures
in the body of the network~\cite{BMR3}.

One can argue that the ability of the network to transform the matrix defining the topology and
weights of this network plays a fundamental role in the context of programming with linear streams, similar to the role of 
$\lambda$-calculus in the context of programming via string rewriting~\cite{BMR4}.

\section{Dataflow Matrix Machines Based on the Vector Space
Generated by Finite Strings}\label{sec:core}

The powerful setup described above involves relatively high level of design complexity.
There are many kinds of linear streams, there are many types of neurons, each neuron type
has its own input and output arity, and a particular kind of linear streams is associated with
each of its inputs and outputs.

This is quite normal in the world of typed programming languages, but it is inconvenient for
Lisp-based frameworks. It also feels more biorealistic not to have strong constraints
and to be able to sculpt and restructure the networks on the fly at runtime, and to run those networks
without fear of runtime exceptions.

It turns out that one can build a setup of sufficient generality based on a single vector
space, and that moreover this vector space is expressive enough to represent activation
functions of variable input and output arities via transformations having one input and one output.

\subsection{Vector Space $V$}

\subsubsection{Finite Linear Combinations of Finite Strings}
Consider a countable set $L$ of tokens (pragmatically speaking, $L$ is often the set of all legal
keys of hash dictionaries in a given programming language). Consider the set $L^*$ of finite sequences
of non-negative length of elements of $L$.

The vector space $V$ is  constructed as the space of finite formal linear combinations of
elements of $L^*$ over reals.

There are several fruitful ways to view elements of $V$. 

\subsubsection{Finite Prefix Trees with Numerical Leaves}

One can associate term
$\alpha l_1 \ldots l_n$ ($\alpha \in \mathbb{R},  l_1 \ldots l_n \in L$),  with a path in a tree with the nodes labeled with $l_1, \ldots, l_n, \alpha$.
Then an element of $V$ (a finite sum of such terms) is associated with a finite
tree with intermediate nodes labeled by elements of $L$, and the leaves being
real numbers. The structure of intermediate nodes is a prefix tree (trie), and
the numerical leaves indicate which paths are actually present, and with what
values of coefficients. 

\subsubsection{``Tensors of Mixed Rank"}
Another way to view an element of $V$ is to associate the empty string (path) with non-zero coefficient (if present in our linear combination)
with a scalar, each string (path) of length one with non-zero coefficient, $\alpha l$,
with the coordinate of a sparse array labeled $l$ taking value $\alpha$, each string (path) of
length two with non-zero coefficient, $\beta l_1 l_2$, with the element of a sparse matrix with the row labeled $l_1$,
the column labeled $l_2$, and the element taking value $\beta$, each string (path)
of length three with non-zero coefficient, $\gamma l_1 l_2 l_3$, with an element of a sparse
``tensor of rank 3",\footnote{When we say 
``tensor of rank $N$", we mean simply a multidimensional array with $N$ dimensions using standard terminology adopted in machine learning.}
etc. 

Therefore, an element of $V$ can in general be considered to be a ``mixed rank" tensor,
a sum of a scalar, a one-dimensional array,  a two-dimensional matrix, a tensor of rank 3, etc.
Moreover, because $L$ is countable, the one-dimensional array in question has countable
number of coordinates, the two-dimensional matrix in question has countable number of
rows and countable number of columns, etc. However, because an element of $V$ is a finite
sum of terms $\alpha l_1 \ldots l_n$, only a finite number of those coordinates are
actually non-zero, and for a given nonzero element of $V$ there is the maximal number $N$ for which its tensor component
of rank $N$ has a non-zero coefficient.

In particular, this means that any usual tensor of a fixed finite shape is representable as
an element of $V$. Therefore, $V$ covers a wide range of situations of interest.
See Appendix~\ref{sec:linear} for a discussion of situations where even higher degree of generality
is needed.

\subsubsection{Recurrent Maps}
One can also represent elements of $V$ via recurrent maps. An element
of $V$ is a pair consisting of a real scalar and a map from $L$ to $V$. The scalar in question
is non-zero if the element of $V$ in question contains the empty string (path)
with non-zero coefficient.

Only a finite number of elements of $L$ can be mapped to non-zero elements of $V$.
One considers the representation of the element of $V$ in question as a finite
prefix tree, and maps elements of $L$ which label the first level of that tree
to the associated subtrees. Other elements of $L$ are mapped to zero.

This representation plays a particularly important role for us. On one hand,
this is the representation which is used to build our current prototype
system~\cite{DMM} in Clojure. In our current implementation $L$ is slightly less
than all legal hash keys in Clojure, namely several keys are reserved for
other purposes. In particular, when the scalar component of an element
of $V$ is non-zero, we simply map the reserved key {\tt :number} to that
value. Therefore, an element of $V$ can always be represented simply
as a Clojure hash map.

Another particularly important use of the recurrent map representation is that the
labels at the first level of a recurrent map can be dedicated to
naming input or output arguments of a function. This is the mechanism to
represent functions of arbitrary input and output arity, and even
functions of variable arity (variadic functions) as functions having one
input and one output. We use this mechanism in the next subsection.

\subsection{DMMs with Variadic Neurons}\label{sec:variadic}

The  activation functions of the neurons
transform single streams of elements of $V$. The labels at the first level
of the elements of $V$ serve as names of inputs and outputs.

The network matrix should provide
a linear transformation mapping all outputs of all neurons to all inputs of all neurons.
Let's consider one input of one neuron, and the row of the network matrix
responsible for computing that input from all outputs of all neurons.
The natural index structure of this row is not flat, but hierarchical. At the very least,
there are two levels of hierarchy: neurons and their outputs. 

In our current implementation
we actually use three levels of hierarchy: neuron types (which are Clojure {\em vars} referring to implementations of
activation functions $V \rightarrow V$), neuron names, and names of the outputs. Therefore, in our current implementation matrix rows are three-dimensional sparse arrays (``sparse tensors of rank 3").

Similarly, the natural index structure for the array of rows is not flat,
but hierarchical. At the very least, there are again two levels of hierarchy:
neurons and their inputs. In our current implementation
we actually use three levels of hierarchy: neuron types, neuron names, and
names of the inputs.

Therefore, in our current implementation the network matrix is a six-dimensional
sparse array (``sparse tensor of rank 6").

Conceptually, the network is countably-sized, but since the network matrix
has only a finite number of non-zero elements at any given time, and hence elements
of $V$ have only a finite number of non-zero coordinates at any given
time, we are always working with finite representations.

On the ``down movement", the network matrix $(w_{f, n_f, i, g, n_g, o}^t)$ (``sparse tensor of rank 6")
is applied to an element of $V$ representing all outputs of all neurons. The result
is an element of $V$ representing all inputs of all neurons to be used during
the next ``up movement".

Here is the formula used to compute one of those inputs:

$$x_{f, n_f, i}^{t+1} = \sum_{g \in F} \sum_{n_g \in L} \sum_{o \in L} w_{f, n_f, i, g, n_g, o}^t * y_{g, n_g, o}^t.$$

Here $f$ and $g$ belongs to the set of neurons types $F$, which is simply the set of
transformations of $V$. Potentially, one can have countable number of
such transformations implemented, but at any given time only finite number
of them are defined and used. The $n_f$ and $n_g$ are names of input and output neurons,
and $i$ and $o$ are the names of the respective input and output arguments of those neurons.

In the formula above, $w_{f, n_f, i, g, n_g, o}^t$ is a number, and $x_{f, n_f, i}^{t+1}$ and $y_{g, n_g, o}^t$ are elements of $V$.

This operation is performed for all  $f \in F$, all $n_f \in L$, all input names $i \in L$ for which
the matrix row has some non-zero elements.

The result is finitely sized map $\{f \mapsto \{n_f \mapsto x_{f, n_f}^{t+1}\}\}$ and each $x_{f, n_f}^{t+1}$
is a finitely sized map from the names of neuron inputs to the values of those inputs, $\{i \mapsto x_{f, n_f, i}^{t+1}\}$.

On the ``up movement" each $f$ is simply applied to the elements of $V$ representing the single inputs
of the activation function $f$ for all the neurons $n_f$ which are
present in this map:

$$y_{f, n_f}^{t+1} = f(x_{f, n_f}^{t+1}).$$

This mechanism is currently used in our implementation of core primitives of dataflow matrix machines
in Clojure~\cite{DMM}. The network matrix $(w_{f, n_f, i, g, n_g, o}^t)$ is obtained as
the output of the {\tt Self} neuron, which adds its two arguments together. The output of {\tt Self} is connected
to one of those inputs with weight 1, making {\tt Self} an accumulator, and {\tt Self} takes additive updates to the network matrix
on its other input, while the network is running. For an example of a similar use of
the {\tt Self} neuron see Appendix~\ref{sec:example}.

\section{DMMs and Machine Learning}\label{sec:conclusion}

There are different ways to view relationships between dataflow matrix machines
and RNNs. One can view DMMs simply as a very powerful generalization of RNNs.
Alternatively, one can view DMMs as a bridge between RNNs and programming languages.

There is already a strong trend to build neural networks from layers and modules rather
than building them from single neurons. For example,  RNN-related classes in TensorFlow~\cite{TensorFlow}
provide strong evidence of that trend. At the same time, engineers looking to
implement and train networks with sparse connectivity patterns or with neurons having multiple inputs or
multiple outputs within TensorFlow framework are well aware that this is a 
much more difficult undertaking, despite appearance of sparse tensors and
activation functions with multiple outputs in TensorFlow documentation.
DMMs encourage us to look at the neural nets with  sufficient degree of generality,
and single DMM neurons can be made powerful enough to serve as layers and modules
when necessary.

In recent years, some authors suggested that synthesis of small functional programs
and synthesis of neural network topology from small number of modules are
closely related problems~\cite{Olah},~\cite{Nejati}. Recently we are seeing progress along each of these
directions (e.g.~\cite{Feser},~\cite{Miikkulainen}). DMMs might provide the right degree of generality to look at these
related classes of problems.

There is strong evidence that syntactic shape of programs and their functionality carry sufficient
mutual information about each other for that to be useful in machine learning inference (e.g.~\cite{SketchLearning}).
Therefore, thinking somewhat more long-term, if DMMs turn out to be a sufficiently popular platform to
handcraft DMM-based software manually, this might provide a corpus of data useful for program synthesis,
similarly to the use of a corpus of hand-crafted code in~\cite{SketchLearning}, potentially giving this approach
an advantage over synthesis of low-level neural algorithms.

The availability of self-referential and self-modifying facilities might be quite attractive from the viewpoint of machine
learning, given their potential for learning to learn and for the network to learn to modify itself,
especially in the context of large networks which continue to gain experience during their lifetime (such as, for example, PathNet~\cite{PathNet}).

One should note that the best learning to learn methods are often those which generalize to a large class of problems~\cite{Wichrowska}.
So the use of self-referential facilities for learning to learn might work better when the network is trained to solve
a sufficiently diverse class of problems, compared to the cases of learning narrow functionality.

\appendices

\section{}\label{sec:linear}

\subsection{Linear Streams of Probabilistic Samples}

In the present paper, we consider DMMs over real numbers.

Sometimes, one needs to represent a stream of large vectors, e.g.
a stream of probability distributions over some measurable space $X$.
One would typically have to approximate such a stream by a stream of samples
drawn from those probability distributions.

In order to have a vector space and to allow linear combinations with negative coefficients, 
we consider the space of all finite signed measures over $X$, and we consider samples to be pairs
$\langle x, s\rangle$, where $x \in X$ and $s$ is a flag taking 1 and -1 as values.

Assume that we have streams of finite signed measures over $X$,  $\mu_1, \ldots, \mu_n$,
and streams of corresponding samples, $\langle x_1, s_1 \rangle, \ldots, \langle x_n, s_n \rangle$.

Let us describe the procedure of computing a sample representing a signed measure
$\alpha_1 * \mu_1 + \ldots + \alpha_n * \mu_n$. We pick index $i$ with
probability $\mid\alpha_i\mid / \sum_j \mid\alpha_j\mid$ and we pick
the sample $\langle x_i, \mbox{sign}(\alpha_i) * s_i \rangle$ to represent
$\alpha_1 * \mu_1 + \ldots + \alpha_n * \mu_n$.

\subsection{Missing Samples and Zero Measures}

The formula in the previous subsection does not work, if all $\alpha_i$
are zero. In general, it is convenient to allow to provide less than one
sample per unit of time, i.e. to allow ``missing samples".

We don't have a complete theory of this situation, which is under development at~\cite{DMM}.
\footnote{\tiny \url{https://github.com/jsa-aerial/DMM/blob/master/design-notes/Early-2017/sampling-formalism.md}}

But at the very least, we do allow missing samples, and we require that when
one is trying to sample from zero measure the result should be the
missing sample.

In particular, if while computing $\alpha_1 * \mu_1 + \ldots + \alpha_n * \mu_n$,
the index $i$ has been picked, and the measure $\mu_i$ is represented by
the missing sample, then the linear combination in question is represented
by the missing sample.

\subsection{Extending Space $V$ to Represent Samples}

The expressive power of space $V$ is insufficient to accommodate streams of
samples from measures. 

The natural generalization in this case is to consider the space of finite
prefix trees with leaves from $\mathbb{R}\oplus M$ instead of $\mathbb{R}$,
where $M$ is the space of signed measures over $X$.

What this would mean implementation-wise is that we are to introduce another
reserved keyword, {\tt :sample}, and a non-zero leaf can contain
{\tt :number {\em numeric-value}}, or {\tt :sample {\em value}}, or both.

The association between missing samples and zero measures fits the
general spirit of space $V$ that zero coordinates should be omitted
from the representations of its elements.

This development is planned for a future version of~\cite{DMM}.

\section{}\label{sec:selfref}

\subsection{Lightweight Pure Dataflow Matrix Machines}

The {\em lightweight} machines use network matrices of finite fixed size instead of the theoretically prescribed
countable-sized matrices with finite number of non-zero elements (for a similar construction see Appendix D of \cite{BMR4}).
Sometimes, it is methodologically convenient to consider this restricted degree of generality.

We consider rectangular matrices $M \times N$.
We consider discrete time, $t = 0, 1, \ldots$, and we consider $M+N$ streams of those rectangular matrices, $X^1, \ldots,
X^M, Y^1, \ldots, Y^N$. At any moment $t$, each of these streams takes a rectangular matrix  $M \times N$ as its value.
(For example, $X^1_t$ or $Y^N_t$ are such rectangular matrices. Elements of matrices are real numbers.)

Let's describe the rules of the dynamical system which would allow to compute $X^1_{t+1}, \ldots, X^M_{t+1}, Y^1_{t+1}, \ldots, Y^N_{t+1}$ 
from $X^1_t, \ldots, X^M_t, Y^1_t, \ldots, Y^N_t$. We need to make a choice, whether to start with $X^1_0, \ldots, X^M_0$ as initial
data, or whether to start with $Y^1_0, \ldots, Y^N_0$. Our equations will slightly depend on this choice. The literature on dataflow matrix machines tends to
start with matrices $Y^1_0, \ldots, Y^N_0$, and so we keep this choice here, even though this might be slightly unusual to the reader. But it is
easy to modify the equations to start with matrices $X^1_0, \ldots, X^M_0$.

Matrix $Y^1_t$ will play a special role, so at any given moment $t$, we also denote this matrix as $A$, and its elements as $a_{i,j}$.
Define $X^i_{t+1} = \sum_{j=1, \ldots, N} a_{i,j} Y^j_t$ for all $i = 1, \ldots, M$. Here $a_{i,j} Y^j_t$ is a matrix resulting from
mutliplying the matrix $Y^j_t$ by number $a_{i,j}$.
Define $Y^j_{t+1} = f^j(X^1_{t+1}, \ldots, X^M_{t+1})$ for all $j=1, \ldots, N$. 

So, $Y^1_{t+1} = f^1(X^1_{t+1}, \ldots, X^M_{t+1})$ defines $Y^1_{t+1}$ which will be used as $A$ at the next time step $t+1$. This is
how the dynamical system modifies itself in lightweight pure dataflow matrix machines.

\subsection{Example of a Self-Modifying Lightweight Pure Dataflow Matrix Machine}\label{sec:example}

This is an example similar to the one from Appendix D.2.2 of \cite{BMR4}. A similar schema is implemented in~\cite{DMM} as\\ 
{\tiny \url{https://github.com/jsa-aerial/DMM/blob/master/examples/dmm/oct_19_2016_experiment.clj}}

Define $f^1(X^1_t, \ldots, X^M_t) = X^1_t + X^2_t$. Start with $Y^1_0 = A$, such that $a_{1,1} = 1$, $a_{1,j} = 0$ for all other $j$,
and maintain the condition that first rows of all other matrices $Y^j, j \neq 1$ are zero. These first rows of all $Y^j, j = 1, \ldots, N$ will be 
invariant as $t$ increases. This condition means that $X^1_{t+1} = Y^1_t$ for all $t \geq 0$.

Let's make an example with 3 constant update matrices: $Y^2_t, Y^ 3_t, Y^4_t$. Namely, say that $f^2(X^1_t, \ldots, X^M_t) = U^2,
f^3(X^1_t, \ldots, X^M_t) = U^3, f^4(X^1_t, \ldots, X^M_t) = U^4$. Then say that $u^2_{2,2} = u^3_{2,3} = u^4_{2,4} = -1$,
and $u^2_{2,3} = u^3_{2,4} = u^4_{2,2} =1$, and that all other elements of $U^2, U^3, U^4$ are zero\footnote{Essentially
we are saying that those matrices ``point to themselves with weight -1", and that ``$U^2$ points to $U^3$, $U^3$ points to $U^4$, and
$U^4$ points to $U^2$ with weight 1".}.
And imposing an additional starting condition on $Y^1_0 = A$, let's say that $a_{2,2} = 1$ and that $a_{2,j} = 0$ for $j \neq 2$.

Now, if we run this dynamical system, the initial condition on second row of $A$ would imply that at the $t=0$, $X^2_{t+1} = U^2$.
Also $Y^1_{t+1} = X^1_{t+1} + X^2_{t+1}$, hence now taking $A = Y^1_1$ (instead of  $A = Y^1_0$), we obtain $a_{2,2} = 1+u^2_{2,2} = 0$,
and in fact $a_{2,j} = 0$ for all $j \neq 3$, but $a_{2,3} = u^2_{2,3} = 1$.

Continuing in this fashion, one obtains $X^2_1 = U^2, X^2_2 = U^3, X^2_3 = U^4, X^2_4 = U^2, X^2_5 = U^3, X^2_6 = U^4,
X^2_7 = U^2, X^2_8 = U^3, X^2_9 = U^4, \ldots$, while the invariant that the second row of matrix $Y^1_t$ has exactly one
element valued at 1 and all other zeros is maintained, and the position of that 1 in the second row of matrix $Y^1_t$ is
2 at $t=0$, 3 at $t=1$, 4 at $t=2$, 2 at $t=3$, 3 at $t=4$, 4 at $t=5$, 2 at $t=6$, 3 at $t=7$, 4 at $t=8$, \ldots

This element 1 moving along the second row of the network matrix is a simple example of a circular wave pattern in the matrix $A = Y^1_t$ controlling the dynamical system in question.

It is easy to use other rows of matrices $U^2, U^3, U^4$ as ``payload" to be placed into the network matrix $Y^1_t$
for exactly one step at a time, and one can do other interesting things with this class of dynamical systems.

\section{}

\subsection{``Two-stroke engine" for a standard DMM}\label{sec:twostrokedmmold}

\begin{figure}[h]
\begin{tikzpicture}
   \clip (-3.5, -2.0) rectangle (7.0, 2.0);

  \filldraw (-3.2,0) circle [radius=0.5pt]
                (-3.0,0)  circle [radius=0.5pt]
                (-3.4, 0) circle [radius=0.5pt]; 

   \draw [->] (-1.2, 0) -- (-0.8, 1) node [right] {$y_{2,C_1}$};
   \draw [->] (-1.2, 0) -- (-1.6, 1) node [left] {$y_{1,C_1}$};
   \draw (-2.0, -1)  node [left] {$x_{1,C_1}$} -- (-1.2, 0);
   \draw (-1.2, -1)  node [below] {$x_{2,C_1}$} -- (-1.2, 0) node [left] {$f_{C_1}$};
   \draw (-0.4, -1)  node [right] {$x_{3,C_1}$} -- (-1.2, 0);

  \filldraw (0,0) circle [radius=0.5pt]
                (-0.2,0)  circle [radius=0.5pt]
                (0.2, 0) circle [radius=0.5pt];

   \draw (1.6, -1) node [left] {$x_{1,C_2}$} -- (2.0, 0)  node [left] {$f_{C_2}$};;
   \draw (2.4, -1) node [right] {$x_{2,C_2}$} -- (2.0, 0);
   \draw [->] (2.0, 0) -- (1.2, 1) node [left] {$y_{1,C_2}$};
   \draw [->] (2.0, 0) -- (2.0, 1) node [above] {$y_{2,C_2}$};
   \draw [->] (2.0, 0) -- (2.8, 1) node [right] {$y_{3,C_2}$};

  \filldraw (3.2,0) circle [radius=0.5pt]
                (3.0,0)  circle [radius=0.5pt]
                (3.4, 0) circle [radius=0.5pt];

  \draw [->, very thick] (0.5, 1.2) .. controls (5.5, 4.2) and (5.5, -4.2) .. (0.5, -1.2)  node [midway, right] {{\bf W}};

\end{tikzpicture}

\caption{``Two-stroke engine" for a standard DMM~\cite{BMR2}.} 
\label{fig:dmmold}
\end{figure}
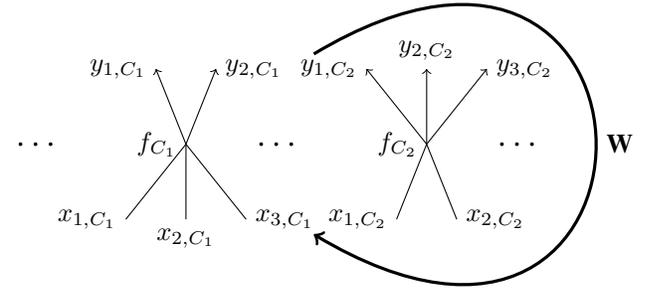

``Down movement": 
for all inputs $x_{i,C_k}$ such that there is a non-zero weight $w_{(i,C_k), (j,C_l)}^t$:
$$x_{i,C_k}^{t+1} = \sum_{\{(j,C_l) | w_{(i,C_k), (j,C_l)}^t \neq 0\}}w_{(i,C_k), (j,C_l)}^t * y_{j, C_l}^{t}.$$ Note that
$x_{i,C_k}^{t+1}$ and $y_{j, C_l}^t$ are no longer numbers, but vectors\footnote{In this Appendix, the formulas are written in terms
of vectors themselves, and not in terms of their approximate representations actually used by the DMM in question.}, so the type correctness condition states that
$w_{(i,C_k), (j,C_l)}^t$ can be non-zero only if $x_{i,C_k}$ and $y_{j, C_l}$ belong to the same vector space.

``Up movement": for all active neurons $C$: $$y^{t+1}_{1,C},...,y^{t+1}_{n_C,C} = f_C (x^{t+1}_{1,C},...,x^{t+1}_{m_C,C}).$$

Because input and output arities are allowed to be zero, special handling of network inputs and outputs which has been
required for RNNs is not required here.

When a formalisms of DMMs based on a single kind of linear streams is used (e.g. DMMs based on streams of matrices in~\cite{BMR4} or DMMs based on streams of finite prefix trees with numerical leaves in Section~\ref{sec:core} of the present paper), the
need for the type correctness condition is eliminated.

\end{document}